\documentclass{article}
\usepackage{amsmath,amssymb,graphicx,mlspconf}
\usepackage{booktabs}
\usepackage{xcolor}

\copyrightnotice{979-8-3195-0884-3/26/\$31.00 {\copyright}2026 IEEE}
\toappear{2026 IEEE International Workshop on Machine Learning for Signal Processing, Sep.\ 28-- Oct.\ 1, 2026, Atlanta, USA}

\title{Physics-Augmented Graph Transformers for Patch-Antenna Forward and Inverse Design}

\name{%
    Avi Epstein
    \qquad Snir Nehemia
    \qquad Haim Suchowski
    \qquad Lior Wolf%
}
\address{%
    Tel Aviv University%
}

\begin{document}
\ninept
\maketitle

\begin{abstract}
Full-wave electromagnetic (EM) simulation enables accurate patch-antenna analysis but is computationally expensive for large-scale forward prediction and inverse design. We present a mesh-native, physics-augmented graph-learning framework that treats radiation-pattern prediction as signal reconstruction on an irregular surface mesh. For the forward problem, a GPS graph transformer is trained with Physics-Augmented Intermediate Supervision (PAIS), an auxiliary node-level objective that predicts complex surface currents, the physical intermediate linking geometry to radiation. PAIS improves multiple GNN backbones at no inference-time cost, while shuffled-current and non-physical controls show the gain comes from physical correspondence. Direction-conditioned decoding and a differentiable radiation-integral consistency loss further exploit this structure. On an 80,000-sample CST benchmark, GPS+PAIS reaches MSE 0.17 / PSNR 19.67, generalizes to a PCA split, and transfers zero-shot to canonical patches. For inverse design, surrogate-filtered diffusion beats nearest-neighbor retrieval by 32\% relative MSE.

\end{abstract}

\section{Introduction}
\label{sec:intro}
Patch antennas are among the most widely deployed radiating structures in modern wireless systems, but designing them remains slow and largely manual. Electromagnetic radiation-pattern prediction can be viewed as a structured signal-reconstruction problem: a continuous angular field must be inferred from a geometric object defined on an irregular spatial domain. While analytical models exist for a limited set of simplified canonical geometries, realistic structures with complex shapes, parasitic elements, and nontrivial boundary conditions require full-wave electromagnetic (EM) simulation. The forward problem, predicting the far-field pattern of a given geometry, is conventionally solved using numerical EM solvers such as CST Microwave Studio \cite{CST2024}, which discretize Maxwell's equations on fine meshes; a single simulation may take minutes to hours. The inverse problem, finding a geometry that produces a desired pattern, is even more demanding, typically requiring dozens to hundreds of iterative EM simulations within an optimization loop.

Replacing this iterative simulation loop with a learned surrogate is a natural target for machine learning. However, progress has been limited by two central challenges. First, antenna geometries are inherently irregular and are poorly represented by regular grid parameterizations commonly used in deep learning. Second, conventional surrogate models trained only on geometry-to-radiation input-output pairs do not explicitly encode the induced current distributions that causally govern far-field radiation.

In this paper, we introduce a physics-augmented machine-learning framework for electromagnetic forward prediction and inverse antenna design. Our approach is motivated by the well-known electromagnetic relationship between induced surface currents and far-field radiation, schematically Geometry $\rightarrow \mathbf{J(r)} \rightarrow$ Far-field: the induced surface-current distribution $\mathbf{J(r)}$ determines the radiated field through the radiation integral (Sec.~\ref{sec:fwd}). We leverage this structure as an intermediate supervision signal during training.

For the forward problem, we develop a mesh-native graph transformer surrogate based on the General, Powerful, Scalable (GPS) transformer \cite{rampavsek2022recipe} operating directly on the antenna surface mesh. It is trained with Physics-Augmented Intermediate Supervision (PAIS), where an auxiliary node-level objective predicts the complex surface-current distribution alongside the far-field pattern. By constraining latent representations to encode this physically meaningful quantity, PAIS consistently improves radiation-pattern prediction across all evaluated graph neural network architectures.

For the inverse problem, we combine a conditional diffusion model with the trained surrogate in a surrogate-filtered generation pipeline: the diffusion model proposes candidate geometries conditioned on a target pattern, and the surrogate rapidly ranks them before final EM validation, sharply reducing the number of expensive full-wave simulations.

Experimentally, the framework achieves strong forward accuracy and robust generalization on a benchmark of 80,000 CST-simulated patch antennas with radiation-pattern and surface-current ground truth. Because this benchmark is generated from FMNIST\slash CIFAR-derived masks, we additionally test transfer beyond the image-derived prior by evaluating GPS\allowbreak+PAIS zero-shot on canonical textbook-style square, rectangular, and parasitic patch antennas not drawn from FMNIST or CIFAR. The surrogate remains accurate without fine-tuning, indicating it is not merely memorizing image-like shape statistics. The inverse pipeline outperforms nearest-neighbor retrieval and surrogate-guided search baselines on the hardest PCA-split targets.

Our contributions are: (i) Physics-Augmented Intermediate Supervision (PAIS), a surface-current auxiliary objective that improves radiation-pattern prediction across graph backbones while adding no inference-time cost, with shuffled-current and non-physical controls showing that the gain comes from physical correspondence rather than generic regularization; (ii) a GPS-based surface-mesh surrogate for patch antennas, augmented with a direction-conditioned decoder and a differentiable radiation-integral consistency loss, providing a current-aware alternative to global graph pooling; (iii) a surrogate-filtered conditional diffusion pipeline for inverse design, in which many generated candidates are ranked cheaply by the forward model and only the top designs are validated in CST; and (iv) an 80,000-sample CST-simulated benchmark with far-field and surface-current ground truth, including PCA-split evaluation and zero-shot validation on textbook-style patch antennas.

\section{Related work}
\label{sec:relwork}

\noindent\textbf{Mesh-based and graph-signal learning.} Graph neural networks are well suited to signals defined on irregular meshes: MeshGraphNets \cite{pfaff2020learning} and graph network simulators \cite{sanchez2020learning} replaced finite-element solvers for fluids and structural mechanics by operating on the native mesh with relative-geometry edge features. Graph transformers such as GPS \cite{rampavsek2022recipe} add global self-attention to local message passing, which matters here because radiation is a global functional of currents across the whole structure. We use this backbone to reconstruct an angular signal from a graph-supported geometry.

\noindent\textbf{EM surrogates and inverse problems.} Data-driven EM surrogates predict near-fields or S-parameters from pixelized geometry \cite{khan2024deep,sarker2023applications} or use hypernetworks for arrays \cite{lutati2021hyperhypernetwork}; physics is sometimes embedded via PDE-residual losses (PINNs \cite{raissi2019physics}) or operator learning (FNO \cite{li2020fourier}, DeepONet \cite{lu2021learning}). We instead supervise $\mathbf{J}$, available as a simulation output and tied to the target through the radiation integral, directly as a structured auxiliary signal - related to graph residual learning for integral-equation solvers \cite{shan2023solving} - and close the loop with a differentiable analytic forward model rather than a PDE-residual penalty. Inverse antenna and metamaterial design, an ill-posed reconstruction problem, has been addressed with VAEs \cite{li2024design}, GANs \cite{al2024generative}, and diffusion \cite{hen2025inverse}; we adopt conditional diffusion with classifier-free guidance \cite{ho2020denoising,ho2022classifier} as a candidate generator filtered by our surrogate.

\section{Dataset}
\label{sec:data}

We introduce a reusable benchmark by combining Fashion-MNIST (FMNIST) \cite{xiao2017fashion} and CIFAR \cite{krizhevsky2009cifar} with full-wave EM simulation in CST. The goal of this benchmark is not to reproduce a library of hand-designed commercial antennas, but to provide a controlled, diverse, reproducible distribution of physically valid radiating geometries. FMNIST/CIFAR-derived masks give a familiar source of structured geometric variation while the EM simulator ensures that all labels are physically generated.

\begin{figure}[t]
\centering
\includegraphics[width=0.94\linewidth]{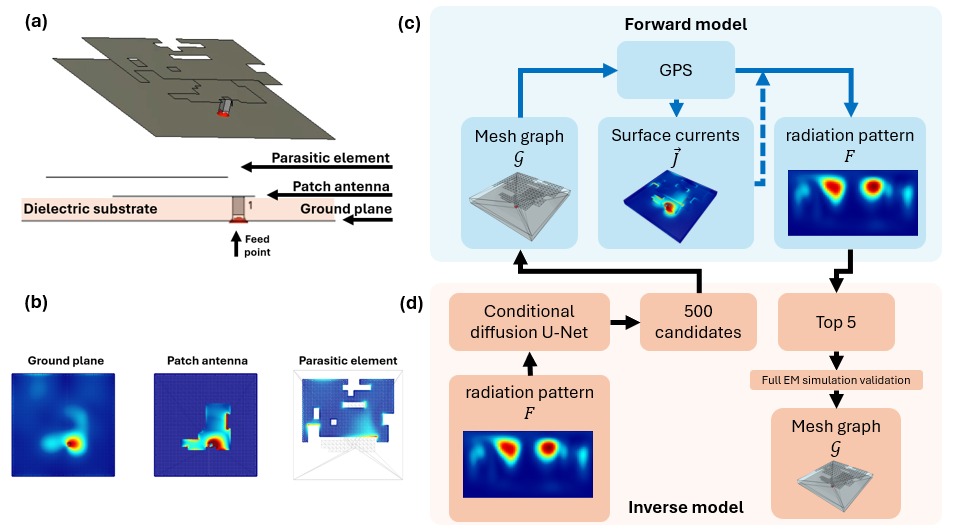}
\caption{Antenna learning and inverse-design pipeline.
(a) Example geometry; (b) component-wise surface currents.
(c) Forward: mesh graph \(G\) $\to$ GPS with surface-current supervision \(\vec{J}\) $\to$ pattern \(F\).
(d) Inverse: target \(F\) conditions a diffusion U-Net; candidates are ranked by the surrogate and validated in CST to select \(G\).}
\label{fig:geom}
\end{figure}

\noindent\textbf{Geometry construction.} Each antenna is parameterized by two binary occupancy matrices: a \emph{patch} matrix obtained by resizing an FMNIST image to $16\!\times\!16$ and binarizing, and a \emph{parasitic element} matrix obtained identically from a CIFAR image. The 3D structure stacks, from top to bottom: a PEC parasitic layer (CIFAR), the PEC patch layer (FMNIST), a PEC feed column placed at the location of the maximum activation in the patch matrix, and a constant PEC ground plane. The gap between the feed column and the ground plane acts as the physical feed gap. This automatic feed placement guarantees every geometry is physically excitable. Representative examples are shown in Fig.~\ref{fig:geom}.

All structures are simulated in CST Studio at 5.6\,GHz using the time-domain solver with open boundary conditions. We release three sub-corpora of approximately 26{,}000 simulated samples each: (i)~\emph{FMNIST-only}, (ii)~\emph{FMNIST+CIFAR}, and (iii)~\emph{random-pixel}, which replaces the FMNIST patch with i.i.d.\ random binary images and expands the geometric distribution beyond structured image priors. Their union (80{,}000 samples) forms the \emph{big dataset} used in our scaling experiments.

\noindent\textbf{Mesh and graph representation.} For graph-based models the surface mesh is used directly as the graph $\mathcal{G}=(\mathcal{V},\mathcal{E})$. Each node carries 3D position, surface normal, a one-hot encoding of structural component type (PEC, Dielectric, FEED), and 10-dimensional Laplacian eigenvector positional encoding (LapPE, $k{=}10$) \cite{dwivedi2021graph}. Each undirected edge is attributed with the relative position vector between its endpoints and its Euclidean norm.

\noindent\textbf{Electromagnetic targets.} For each sample, CST stores two outputs. The \emph{radiation pattern} is recorded over the elevation-azimuth sphere at $34\!\times\!34$ resolution ($\theta\!\times\!\phi$) in linear gain normalized to directivity; this is the primary prediction target. The \emph{surface current distribution} is exported as a point cloud and mapped onto the mesh nodes via $k$-nearest-neighbor interpolation, giving each node a six dimensional vector $(J_x, J_y, J_z)\in\mathbb{C}^3$ stored as real and imaginary parts. The radiation pattern is uniquely determined from $\mathbf{J}$ via the radiation integral; we exploit this causal structure in PAIS (Sec.~\ref{sec:method}).

\noindent\textbf{Splits.} We use two complementary splits. A 90/10 \emph{random split} measures interpolation accuracy. A \emph{PCA split} evaluates generalization to electromagnetically atypical antennas: PCA is applied to the flattened radiation patterns (not to the geometry matrices) and the 10\% of samples with the largest absolute projection onto the first principal component are held out. We further isolate the 100 hardest samples in this PCA split, those farthest from the mean projection as a concentrated stress test for the inverse design.

\section{Method}
\label{sec:method}

We address two complementary problems. The \emph{forward problem} predicts the radiation pattern of a given 3D antenna without running a full-wave simulation. The \emph{inverse problem} synthesizes an antenna geometry that realizes a desired radiation pattern. The forward surrogate is a graph transformer with PAIS (Sec.~\ref{sec:fwd}); the inverse model is a conditional diffusion sampler filtered by that surrogate at inference time (Sec.~\ref{sec:inv}). Fig.~\ref{fig:geom} summarizes both halves.

\subsection{Forward problem: GPS with PAIS}
\label{sec:fwd}

\noindent\textbf{GPS architecture.} Each antenna is the attributed mesh graph from Sec.~\ref{sec:data}, avoiding lossy re-discretization onto a grid. We adopt the GPS graph transformer \cite{rampavsek2022recipe}, which interleaves a GINEConv message-passing sublayer \cite{xu2018powerful,hu2019strategies} with global multi-head self-attention at each block. By default, node embeddings are sum-pooled and decoded by a three-layer MLP to the $34{\times}34$ pattern; we also introduce a direction-conditioned decoder below that removes this pooling step.

\noindent\textbf{Physical motivation.}
For a current distribution $\mathbf{J}(\mathbf{r})$ on a perfectly conducting
surface, the radiated far field in direction
$\hat{\mathbf{k}}(\theta,\phi)$ is given by the radiation integral
\begin{equation}
\mathbf{F}(\theta,\phi) \;\propto\;
\hat{\mathbf{k}}\times\!\!\int_{S}
\mathbf{J}(\mathbf{r})\, e^{\,j k\, \hat{\mathbf{k}}\cdot\mathbf{r}}\, dS,
\label{eq:radint}
\end{equation}
with $k = 2\pi/\lambda$ and
$\hat{\mathbf{k}} = (\sin\theta\cos\phi,\,\sin\theta\sin\phi,\,\cos\theta)$,
and the radiation pattern is $|\mathbf{F}|^2$
normalized to directivity. 

Two properties of Eq.~\eqref{eq:radint} drive our design. First, the map is \emph{causal and factored}: geometry fixes the boundary conditions that induce $\mathbf{J}$, and $\mathbf{J}$ alone - through a known, fixed integral - determines the pattern. The surface current is therefore a physically grounded intermediate through which much of the geometry-to-far-field mapping is mediated. In our setting the dielectric substrate alters the background Green's function, so the field is a substrate-dependent functional of $\mathbf{J}$ rather than the free-space integral of Eq.~\eqref{eq:radint}; we therefore do not use the closed form directly but learn its effective kernel in the decoder. Second, the pattern in each direction is a \emph{phase-weighted linear aggregation} of the \emph{local} currents at every surface point, with direction entering only through the propagation phase $e^{\,j k\, \hat{\mathbf{k}}\cdot\mathbf{r}}$.

\noindent\textbf{Physics-Augmented Intermediate Supervision (PAIS).}
Equation~\eqref{eq:radint} motivates the geometry-to-pattern map as geometry $\rightarrow \mathbf{J} \rightarrow$ far-field, with $\mathbf{J}$
serving as a physically grounded intermediate between the mesh geometry and the radiated pattern. PAIS exploits this structure
with a node-level auxiliary objective: a per-node MLP attached to the node
embeddings \emph{before} pooling predicts the six-dimensional complex
surface current at each mesh node ($J_x, J_y, J_z$, real and imaginary).
Training minimizes
\begin{equation}
\mathcal{L} = \lambda_J \,\lVert \hat{\mathbf{J}} - \mathbf{J} \rVert_2^2
            + \lambda_F \,\lVert \hat{\mathbf{F}} - \mathbf{F} \rVert_2^2
            + \lambda_{\text{phys}}\, w(t)\,
              \lVert \mathbf{F}_{\text{phys}}(\hat{\mathbf{J}}) - \mathbf{F} \rVert_2^2,
\label{eq:loss}
\end{equation}
where the first two MSE terms act on the current components and on the
radiation pattern, and the third is the physics-consistency term introduced
below (Eq.~\eqref{eq:loss} with $\lambda_{\text{phys}}{=}0$ recovers plain
PAIS). Here $\mathbf{F}_{\text{phys}}(\hat{\mathbf{J}})$ is the pattern
obtained by passing the predicted currents through a differentiable
discretization of Eq.~\eqref{eq:radint}, and
$w(t)=\min(1,\,t/T_{\text{warm}})$ with $T_{\text{warm}}{=}2000$ is a linear
warmup that ramps the term in over early training, once the surface-current
head has begun to produce meaningful $\hat{\mathbf{J}}$.
Because current magnitudes span a much larger numerical range than pattern
values ($0$--$40$ vs.\ $0$--$4$), the plain-PAIS rows
($\lambda_{\text{phys}}{=}0$) downweight the current term
($\lambda_J{=}0.1$, $\lambda_F{=}0.9$); the ``+Physical'' configuration
reweights to $\lambda_F{=}0.6$, $\lambda_J{=}0.4$,
$\lambda_{\text{phys}}{=}0.3$.

Node-level supervision helps by a representation-bottleneck argument. The pattern is decoded from a single pooled embedding, and pooling is lossy: a model supervised only on the pattern can discard local information and learn a geometry-to-far-field shortcut. Yet radiation is a global functional of a \emph{local} signal -- the per-node currents -- exactly what pooling washes out; forcing the pre-pooling embeddings to reconstruct $\mathbf{J}$ keeps this local-to-global structure in the representation the decoder consumes. The auxiliary head is training-only, so PAIS adds no inference cost and is architecture-agnostic.

\noindent\textbf{Direction-conditioned decoder.} To remove the pooling bottleneck entirely rather than merely compensating for it, we introduce an alternative decoder that mirrors the radiation integral, whose per-direction field is a weighted sum of contributions from the surface current at every point. We form one query per radiation-pattern direction $(\theta,\phi)$, encoded as a six-dimensional feature $[k_x,k_y,k_z,\sin\theta,\sin\phi,\cos\phi]$ where $\mathbf{k}=(\sin\theta\cos\phi,\allowbreak\sin\theta\sin\phi,\allowbreak\cos\theta)$ is the unit propagation direction and $\phi$ is encoded periodically so the $\pm\pi$ azimuth seam carries no discontinuity. These direction queries cross-attend over the per-node embeddings, and the predicted surface currents are concatenated to the embeddings on both the key and value paths, so that the attention \emph{scores} - not merely the aggregated values - depend on the current at each node. Each $(\theta,\phi)$ output is thus a learned, current-weighted aggregation over mesh nodes, a soft analogue of the radiation integral that consumes the same node-level signal PAIS regularizes. This decoder is the GPS+PAIS+Directional configuration in Table~\ref{tab:fwd}.

\noindent\textbf{Physics-consistency loss via a differentiable radiation
integral.} A third instantiation of the same factorization closes the loop
end-to-end. A differentiable analytic
radiation layer applies Eq.~\eqref{eq:radint} directly over the mesh nodes:
it evaluates the integral as a discrete phase-weighted sum of the per-node
predicted currents $\hat{\mathbf{J}}$ and normalizes to gain/directivity,
yielding a non-learned pattern
$\mathbf{F}_{\text{analytic}}(\hat{\mathbf{J}})$ that is differentiable in
$\hat{\mathbf{J}}$.

The idealized discrete integral omits effects present in the full-wave
labels - meshing-density variation, the ground-plane image, the
dielectric/substrate background Green's-function effect, and the
feed-excitation phase reference. We therefore add a small learned residual: a
lightweight MLP head on the pooled graph embedding produces an additive
correction,
\begin{equation}
\mathbf{F}_{\text{phys}}(\hat{\mathbf{J}}) =
\mathbf{F}_{\text{analytic}}(\hat{\mathbf{J}})
+ s\cdot\text{MLP}(\mathbf{h}),
\qquad s = 0.1,
\label{eq:phys}
\end{equation}
where $\mathbf{h}$ is the pooled graph embedding and $s$ a fixed scale. The
residual absorbs these un-modeled substrate and ground-plane effects without
letting the network bypass the physics, since the analytic term carries the
gradient back to $\hat{\mathbf{J}}$. The result is compared against the
\emph{ground-truth} target $\mathbf{F}$ -- not the model's own
$\hat{\mathbf{F}}$ -- under the same far-field loss, so the predicted currents
must be physically consistent with the target pattern. This is the
$\lambda_{\text{phys}}$ term in Eq.~\eqref{eq:loss}, gated by the warmup
$w(t)$ above; it is the GPS+PAIS+Directional+Physical configuration in
Table~\ref{tab:fwd}. A residual-off ablation ($s{=}0$) isolating the
analytic term is reported in Sec.~\ref{sec:exp_fwd}.

\noindent\textbf{Baselines.} We compare GPS against grid-based models (MLP, U-Net, ResNet-50), other graph-based models (GCN\cite{kipf2016semi}, \cite{velivckovic2017graph}, Mesh Graph Networks\cite{pfaff2020learning}, Graph U-Net\cite{gao2019graph}, DGCNN\cite{wang2019dynamic}), and a nearest-neighbor retrieval baseline. Each GNN baseline is evaluated both with and without PAIS to isolate the contribution of physics-augmented supervision from the choice of architecture.

\subsection{Inverse synthesis via the learned surrogate}
\label{sec:inv}

As an application of the forward surrogate, we use it to filter a generative proposal distribution for inverse synthesis. Each geometry is two $16\!\times\!16$ binary occupancy matrices $\mathbf{x}\in\{0,1\}^{2\times16\times16}$; the feed is placed deterministically at the patch's maximum-activation pixel, so no feed-placement head is needed. A conditional denoising U-Net generates candidates given the target pattern $\mathbf{F}^*$, injected both by concatenating a downsampled $\mathbf{F}^*$ to the noisy geometry and by multi-scale cross-attention in the decoder. We train the standard DDPM \cite{ho2020denoising} $\varepsilon$-objective ($\beta_1{=}10^{-4}$, $\beta_T{=}0.02$, $T{=}700$) with classifier-free guidance \cite{ho2022classifier} ($p_{\text{uncond}}{=}0.1$); at inference we use guidance scale $w{=}1$ (larger $w$ tightens the top-$K$ set and reduces coverage) and binarize samples at $0.5$.

Diffusion is not the central contribution here (PAIS is); it was chosen to produce diverse, target-conditioned proposals for the one-to-many inverse map, and a GAN could serve the same role. Plug-and-play and unrolled solvers typically require gradient or proximal data-fidelity steps \emph{through} the design-to-response map, which is nontrivial across binarization and mask-to-mesh conversion; sample-then-filter needs only forward evaluations. Table~\ref{tab:inv} separates the prior from the filtering effect.

The surrogate then performs best-of-$N$ selection: a single sample need not match the target since the inverse map is one-to-many, so for each $\mathbf{F}^*$ we draw $N{=}500$ candidates, score them with GPS+PAIS in one batched pass, and validate only the top $K{=}5$ in EM simulation. This replaces $500$ full-wave solves per target with $5$ ($100\times$ fewer), plus seconds of GPU scoring. $N{=}500$ is the practical knee: $N{=}100$ worsens inverse MSE by ${\sim}0.08$ on the 100-hardest set, and $N{=}1000$ gives negligible gain.

\section{Experiments}
\label{sec:exp}

\subsection{Setup}
All forward comparisons use the $34\!\times\!34$ radiation pattern with four metrics: MSE, MAE, PSNR, and multi-scale structural similarity (MS-SSIM)\cite{wang2003multiscale}. For inverse design we also report HPBW-IoU, the intersection-over-union between target and predicted angular regions within 0.5 of their respective gain maxima (higher = better main-lobe agreement).
Forward models train for 150 epochs (batch 16, AdamW, base LR $2\!\times\!10^{-4}$, weight decay $10^{-5}$, 5-epoch warm-up, cosine annealing to $10^{-6}$). The diffusion model trains with Adam (LR $10^{-4}$, batch 32, 1000 epochs). All training uses a single NVIDIA RTX 4090.

\noindent\textbf{Computational cost.} GPS+PAIS has 2.6M parameters and trains in ${\sim}12$\,h on one RTX 4090; inference costs 2.6\,GFLOPs / 1.3\,ms per antenna, versus ${\sim}2$--$5$\,min per CST solve (${\sim}10^{5}\times$ speedup). The diffusion U-Net has 3.5M parameters and trains in ${\sim}16$\,h; generating 500 candidates ($T{=}700$, 2 CFG evaluations per step) costs ${\sim}420$\,TFLOPs / ${\sim}48$\,s. The surrogate ranks all 500 candidates, and only the top 5 reach CST: ${\sim}10$--$25$\,min total, versus ${\sim}17$--$42$\,h for validating all 500.

\subsection{Forward problem}
\label{sec:exp_fwd}

Table~\ref{tab:fwd} reports forward results on FMNIST+CIFAR under the random split. Every learned model beats the nearest-neighbor baseline, confirming the task is not solved by memorization. Graph models with PAIS consistently outperform grid baselines of comparable capacity: the MLP, U-Net and ResNet-50 cluster at MSE $\approx 0.53$--$0.56$ / PSNR $\approx 14.0$--$14.2$, while GPS without PAIS reaches MSE~0.45 / PSNR~15.2 and GPS+PAIS reaches MSE~0.39 / PSNR~16.1 / MS-SSIM~0.86. The direction-conditioned decoder (GPS+PAIS+Directional) further lowers error to MSE~0.35 / PSNR~16.35, and the physics-consistency loss (GPS+PAIS+Directional+Physical) gives the best configuration, MSE~0.33 / PSNR~16.78 / MS-SSIM~0.88. Most importantly, \textbf{PAIS improves nearly every GNN architecture, with gains that scale with backbone capacity}: across GCN, GAT, MGN, Graph U-Net and GPS the surface-current head reduces MSE and raises PSNR/MS-SSIM over the same backbone without it, with the largest gains on the highest-capacity backbones (GPS +0.06 MSE / +0.9\,dB; MGN +0.03 MSE / +0.4\,dB); only DGCNN is near-flat. Two controls on GPS, matched in architecture and loss weighting, isolate the cause: replacing the target with node 3D coordinates (``pos.\ aux'') or with currents permuted across the dataset (``shuffled SC'') both collapse to the no-PAIS baseline (MSE ${\approx}$0.45 / PSNR ${\approx}$15.1), while only the correct physical target recovers the full gain (MSE 0.39 / PSNR 16.10). The improvement is therefore attributable to physical correspondence, not the auxiliary objective in isolation.

\noindent\textbf{Residual-off ablation of the physics layer.} With the learned residual disabled ($s{=}0$), the analytic radiation term alone improves over the no-physics configuration (Table~\ref{tab:fwd}: MSE 0.34 vs.\ 0.35, ${\sim}3$\% MSE, +0.25\,dB), and the constrained residual ($s{=}0.1$) improves further to MSE 0.33 / PSNR 16.78. Applying the analytic layer to the \emph{ground-truth} currents yields MSE 1.333, quantifying the aggregate mismatch from substrate, ground-plane, feed-reference, and discretization effects that the residual absorbs.

\begin{table}[t]
\centering
\caption{Forward radiation-pattern prediction on FMNIST+CIFAR (random split, 5.6\,GHz). Lower is better for MAE/MSE; higher for MS-SSIM/PSNR. ``+PAIS'': Physics-Augmented Intermediate Supervision; ``pos. aux''/``shuffled SC'' are controls (node coordinates / permuted currents); ``+Directional'' and ``+Phys'' add the direction-conditioned decoder and physics-consistency loss (Sec.~\ref{sec:fwd}). Best in bold; DGCNN, MGN and GPS rows are means over 3 seeds (std $\le0.01$ for MAE/MSE/MS-SSIM and $\le0.18$\,dB for PSNR); the matched-size mixture row uses one seed.}
\label{tab:fwd}
\renewcommand{\arraystretch}{0.9}
\setlength{\tabcolsep}{5.2pt}
\scriptsize
\begin{tabular}{lcccc}
\toprule
Model & MAE$\downarrow$ & MSE$\downarrow$ & MS-SSIM$\uparrow$ & PSNR$\uparrow$ \\
\midrule
Nearest neighbor   & 0.60 & 1.00 & 0.70 &  9.54 \\
MLP                & 0.45 & 0.55 & 0.81 & 14.11 \\
U-Net              & 0.45 & 0.53 & 0.81 & 14.23 \\
ResNet-50          & 0.46 & 0.56 & 0.80 & 14.00 \\
\midrule
GCN                & 0.44 & 0.53 & 0.81 & 14.46 \\
GCN + PAIS         & 0.43 & 0.50 & 0.82 & 14.63 \\
GAT                & 0.42 & 0.47 & 0.83 & 14.93 \\
GAT + PAIS         & 0.40 & 0.45 & 0.83 & 15.06 \\
Graph U-Net        & 0.45 & 0.54 & 0.81 & 14.20 \\
Graph U-Net + PAIS & 0.42 & 0.48 & 0.82 & 14.80 \\
DGCNN              & 0.45 & 0.54 & 0.80 & 14.19 \\
DGCNN + PAIS       & 0.44 & 0.53 & 0.81 & 14.23 \\
MGN                & 0.42 & 0.47 & 0.83 & 14.83 \\
MGN + PAIS         & 0.40 & 0.44 & 0.83 & 15.25 \\
GPS                & 0.40 & 0.45 & 0.83 & 15.20 \\
GPS + PAIS         & 0.37 & 0.39 & 0.86 & 16.10 \\
GPS + PAIS + Directional & 0.35 & 0.35 & 0.87 & 16.35 \\
GPS + PAIS + Dir.\ + Phys ($s{=}0$) & 0.35 & 0.34 & 0.87 & 16.60 \\
GPS + PAIS + Directional + Phys & \textbf{0.34} & \textbf{0.33} & \textbf{0.88} & \textbf{16.78} \\
\midrule
GPS + shuffled SC (ablation) & 0.40 & 0.45 & 0.83 & 15.12 \\
GPS + pos. aux (ablation)    & 0.41 & 0.45 & 0.84 & 15.14 \\
GPS + PAIS (mixed, matched size) & 0.31 & 0.23 & 0.89 & 19.20 \\
GPS + PAIS (big)             & 0.26 & 0.17 & 0.92 & 19.67 \\
GPS + PAIS (big-PCA split)   & 0.25 & 0.18 & 0.94 & 19.40 \\
\midrule
\multicolumn{5}{l}{\scriptsize\emph{Zero-shot on canonical patches, GPS + PAIS (big) (no fine-tuning)}}\\
Classic square patch  & 0.25 & 0.06 & 0.97 & 24.76 \\
Classic square patch + parasitic & 0.40 & 0.45 & 0.87 & 14.49 \\
Classic rectangular patches & 0.21 & 0.13 & 0.94 & 22.70 \\
\bottomrule
\end{tabular}
\end{table}

\begin{figure}[t]
\centering
\includegraphics[width=0.84\linewidth]{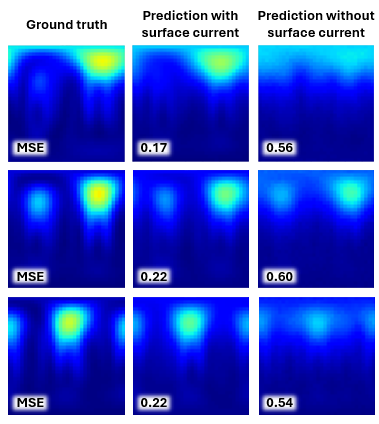}
\caption{\textbf{Zero-shot predictions on square patch antennas with a parasitic element.} Each row: ground-truth pattern (left), prediction with surface-current information (center), and without it (right); MSE is shown lower-left of each prediction.} 
\label{fig:zeroshot_images}
\end{figure}

\noindent\textbf{Zero-shot validation on textbook patch antennas.} To test whether the surrogate generalizes beyond image-derived geometries, we additionally evaluate GPS+PAIS (big) zero-shot on three small held-out sets of realistic, rule-based patch antennas of the kind found in standard antenna-design textbooks: 64 single square-patch designs, 64 patches with a parasitic reflector, and 108 rectangular patches of varying width, each generated by sweeping the patch location and feed-edge position on the $16\times16$ grid (Table~\ref{tab:fwd}, lower block). Without any fine-tuning, the single-patch and rectangular sets are markedly easier than the random-split FMNIST+CIFAR result, consistent with these geometries lying well within the convex hull of training shapes. Adding a parasitic reflector raises the difficulty back to the FMNIST+CIFAR random-split level, indicating that the additional radiating layer, rather than the image-derived geometry distribution, is the dominant source of forward-prediction error. Examples can be seen in Fig.~\ref{fig:zeroshot_images}.

\noindent\textbf{Scaling and OOD generalization.} Retraining GPS+PAIS on the big dataset (80{,}000 samples, Table~\ref{tab:fwd}) reduces MSE by a further 0.22 and improves PSNR by 3.6\,dB, indicating that the architecture is not saturated and benefits substantially from broader geometric coverage. A stratified mixture of the three sub-corpora at the \emph{same} training-set size as FMNIST+CIFAR (``mixed, matched size'', one seed) reaches MSE~0.23 / PSNR~19.2 -- a 41\% MSE reduction at fixed sample count -- so diversity contributes substantially beyond scale; the full 80k set further reduces MSE from 0.23 to 0.17. On the PCA split this configuration attains MSE~0.18, MAE~0.25, MS-SSIM~0.94 and PSNR~19.4 - essentially matching its random split performance, despite the held-out 10\% consisting of the samples electromagnetically furthest from the bulk of the training distribution. This supports the use of GPS+PAIS as the forward surrogate in the inverse pipeline (Sec.~\ref{sec:exp_inv}), where targets may be far from typical patch-antenna radiation patterns. The big-dataset, PCA-split, and zero-shot results use the pooled-decoder GPS+PAIS variant; integrating the direction-conditioned decoder at this scale is left to future work.

\subsection{Inverse problem}
\label{sec:exp_inv}

For each target $\mathbf{F}^*$ the diffusion model generates $N{=}500$ candidates; each is binarized, the feed placed deterministically, and the candidate scored by GPS+PAIS. The top $K{=}5$ by surrogate MSE are validated in CST and we report the best (Diff.\ + surr.). We compare against: nearest-neighbor retrieval (NN); three classical surrogate-guided optimizers (CMA-ES, GA, SA) from random binary init under the same 500-call budget, as a probe of the discrete design space; an NN-seeded SA variant; and Diff.\ -- 5 rand., five unfiltered diffusion samples, isolating surrogate filtering from the generative prior.

Table~\ref{tab:inv} reports the 100 hardest PCA-split targets. The pipeline beats NN by a wide margin -- MSE 0.77 vs.\ 1.13 (32\% relative), PSNR 11.12 vs.\ 9.02\,dB, HPBW-IoU 0.77 vs.\ 0.53. Three comparisons localize the gain. The from-random optimizers all collapse (MSE 2.76--3.30, HPBW-IoU $\le$0.23), so local search over the discrete space fails even with an accurate surrogate. NN-seeded SA recovers most of NN (MSE 1.49) but still trails by 0.72 MSE, so initialization is not the explanation. Diff.\ -- 5 rand.\ (MSE 1.95) sits between the two, confirming that surrogate filtering, not the prior alone, closes and reverses the gap to NN. The same ordering holds on the 500 hardest targets (Table~\ref{tab:inv500}), where NN only edges MS-SSIM (0.88 vs.\ 0.87) because retrieval returns physically realized patterns that stay structurally plausible even when gain values are off. The 0.20 MSE gap between the GNN-eval and simulated rows of Table~\ref{tab:inv} reflects the surrogate's residual error on OOD geometries, consistent with Sec.~\ref{sec:exp_fwd}.

\begin{table}[t]
\centering
\caption{Inverse design on the 100 hardest PCA-split targets. All rows except ``GNN eval.'' are CST-validated against ground truth. GA/SA: genetic algorithm / simulated annealing over the two $16{\times}16$ matrices under the same 500-evaluation budget; NN-seeded SA starts from the nearest training antenna. ``Diff.\ -- 5 rand.'': 5 unfiltered diffusion samples. ``GNN eval.'': the surrogate's own score of the selected candidate.}
\label{tab:inv}
\setlength{\tabcolsep}{3.8pt}
\scriptsize
\begin{tabular}{lccccc}
\toprule
Method & MAE$\downarrow$ & MSE$\downarrow$ & MS-SSIM$\uparrow$ & PSNR$\uparrow$ & HPBW-IoU$\uparrow$ \\
\midrule
NN                      & 0.61 & 1.13 & 0.85 & 9.02  & 0.53 \\
NN-seeded SA + surr.    & 0.59 & 1.49 & 0.83 & 9.13  & 0.52 \\
CMA                     & 0.85 & 2.76 & 0.70 & 6.00  & 0.23 \\ 
GA + surr.              & 0.94 & 3.30 & 0.65 & 4.99  & 0.16 \\
SA + surr.              & 0.95 & 3.09 & 0.65 & 5.08  & 0.18 \\
Diff.\ -- 5 rand.       & 0.70 & 1.95 & 0.79 & 7.45  & 0.36 \\
Diff.\ + surr.\ (GNN eval.) & 0.44 & 0.57 & 0.90 & 12.01 & --   \\
\midrule
\textbf{Diff.\ + surr.} & \textbf{0.49} & \textbf{0.77} & \textbf{0.85} & \textbf{11.12} & \textbf{0.77} \\
\bottomrule
\end{tabular}
\end{table}

\begin{table}[t]
\centering
\caption{Inverse design on the 500 hardest PCA split targets. ``Optimized (simulated)'' is the best of the top-5 surrogate selected candidates, validated in EM simulation.}
\label{tab:inv500}
\setlength{\tabcolsep}{4pt}
\scriptsize
\begin{tabular}{lccccc}
\toprule
Method & MSE$\downarrow$ & MAE$\downarrow$ & PSNR$\uparrow$ & MS-SSIM$\uparrow$ & HPBW-IoU$\uparrow$ \\
\midrule
NN (simulated)        & 0.65 & 0.48 & 11.33 & \textbf{0.88} & 0.68 \\
Optimized (simulated) & \textbf{0.56} & \textbf{0.41} & \textbf{12.74} & 0.87 & \textbf{0.79} \\
\bottomrule
\end{tabular}
\end{table}

\begin{figure}[t]
\centering
\includegraphics[width=0.88\linewidth]{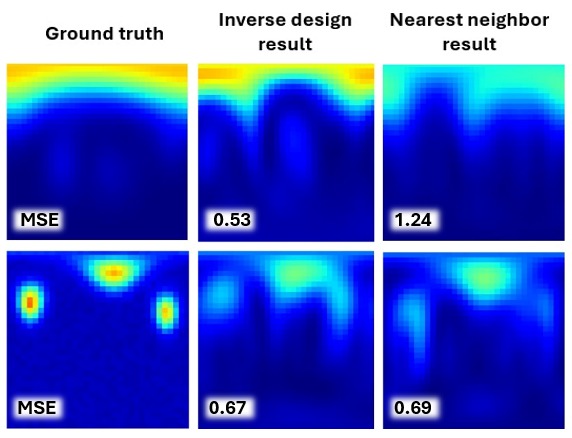}
\caption{\textbf{Hand-designed targets} ($w{=}2$). Columns: surrogate-filtered diffusion output (CST-simulated), target, and nearest-neighbor retrieval. Top: multi-lobe target; bottom: broad-lobe target.} 
\label{fig:probe}
\end{figure}

\noindent\textbf{Qualitative probe on hand-designed patterns.} On two targets outside the image-derived prior - a multi-lobe and a broad single-lobe target with sharp boundaries (Fig.~\ref{fig:probe}) - the optimized designs slightly outperform NN but miss fine local lobes and the asymmetric peak, indicating limited extrapolation rather than general-purpose synthesis; broadening the geometric prior is a natural next step.

\section{Conclusion}
\label{sec:conc}
We presented an end-to-end pipeline for patch-antenna radiation-pattern prediction and inverse design, unified by one observation: the surface-current distribution provides a physically grounded intermediate between geometry and radiated field. We turned this into a supervision signal PAIS that improves nearly every GNN architecture tested, with gains that grow with backbone capacity, vanish under shuffled or non-physical targets, and add no inference cost. A direction-conditioned decoder replacing global pooling with current-weighted cross-attention over mesh nodes further improves random-split accuracy, while a differentiable radiation-integral consistency loss closes the loop between the predicted currents and the radiated field. The surrogate generalizes essentially without loss to a PCA split of the most atypical antennas and transfers zero-shot to canonical textbook geometries (square/rectangular/parasitic MSE 0.06/0.13/0.45), complementing the image-derived benchmark with transferable electromagnetic structure. The inverse pipeline ranks 500 diffusion candidates in seconds, validates five in CST, and beats nearest-neighbor retrieval on the hardest out-of-distribution targets by 32\% relative MSE. We release a benchmark of 80,000 CST-simulated antennas with radiation-pattern and surface-current ground truth. Limitations point to next steps: the geometric prior is image-derived and restricted to binary occupancy masks on a fixed stack -- continuous geometries, material/dielectric variation, and multilayer stacks remain future work; the diffusion sampler struggles on patterns unlike any patch antenna's; the inverse study ranks candidates with the pooled surrogate, and scaling the direction-conditioned decoder remains future work; and the benchmark is single-frequency, making wideband, multi-frequency extension the most direct next direction.


\fontsize{9pt}{9.0pt}\selectfont
\bibliographystyle{IEEEbib}
\bibliography{strings, refs}
\end{document}